\pdfoutput=1

\documentclass[11pt]{article}

\usepackage{acl}

\usepackage{times}
\usepackage{latexsym}

\usepackage[T1]{fontenc}

\usepackage[utf8]{inputenc}

\usepackage{microtype}

\usepackage{amsfonts}
\usepackage{amsmath}
\usepackage{booktabs}
\usepackage{graphicx}
\usepackage{url}
\usepackage[toc,page]{appendix}

\usepackage{longtable}
\usepackage{tabu}
\usepackage{tabularx}
\usepackage{xtab}
\usepackage{balance}
\usepackage{wrapfig}
\usepackage{CJK,algorithm,algorithmic,amssymb,array,epsfig,graphics,multirow,float,subcaption,verbatim,epstopdf}
\usepackage{enumitem}
\usepackage{color,soul} 
\usepackage{hhline}
\usepackage{multirow}
\usepackage{xcolor}
\usepackage{hyperref}
\usepackage{seqsplit}
\usepackage{stmaryrd}
\usepackage{color}
\usepackage{fdsymbol}

\DeclareUnicodeCharacter{03B2}{\ensuremath{\beta}}

\usepackage{xcolor}
\newcommand{\parm}{\mathord{\color{black!33}\bullet}}

\newcount\Comments   
\definecolor{purple}{rgb}{1,0,1}
\newcommand{\kibitz}[2]{\ifnum\Comments=1\textcolor{#1}{#2}\fi}

\usepackage{xparse}
\NewDocumentCommand{\qingyun}{ mO{}}{\textcolor{cyan}{\textsuperscript{\textit{Qingyun}}\textsf{\textbf{\small[#1]}}}}

\title{Named Entity Recognition Under Domain Shift via Metric Learning\\ for Life Sciences}

\author{
  Hongyi Liu$^{1}$, \ Qingyun Wang$^{2}$, \ Payam Karisani$^{2}$, \ Heng Ji$^{2}$  \\
  $^{1}$ Shanghai Jiao Tong University, $^{2}$ University of Illinois at Urbana-Champaign \\
  \texttt{liu.hong.yi@sjtu.edu.cn},\\ \texttt{\{qingyun4,karisani,hengji\}@illinois.edu}\\}
\NewDocumentCommand{\heng}
{ mO{} }{\textcolor{red}{\textsuperscript{\textit{Heng}}\textsf{\textbf{\small[#1]}}}}

\begin{document}
\maketitle

\begin{abstract}
Named entity recognition is a key component of Information Extraction (IE), particularly in scientific domains such as biomedicine and chemistry, where large language models (LLMs), e.g., ChatGPT, fall short. We investigate the applicability of transfer learning for enhancing a named entity recognition model trained in the biomedical domain (the source domain) to be used in the chemical domain (the target domain). A common practice for training such a model in a few-shot learning setting is to pretrain the model on the labeled source data, and then, to finetune it on a hand-full of labeled target examples. In our experiments, we observed that such a model is prone to mislabeling the source entities, which can often appear in the text, as the target entities. To alleviate this problem, we propose a model to transfer the knowledge from the source domain to the target domain, but, at the same time, to project the source entities and target entities into separate regions of the feature space. This diminishes the risk of mislabeling the source entities as the target entities. Our model consists of two stages: 1) entity grouping in the source domain, which incorporates knowledge from annotated events to establish relations between entities, and 2) entity discrimination in the target domain, which relies on pseudo labeling and contrastive learning to enhance discrimination between the entities in the two domains. We conduct our extensive experiments across three source and three target datasets, demonstrating that our method outperforms the baselines by up to 5\% absolute value\footnote{Code, data, and resources are publicly available for research purposes: \url{https://github.com/Lhtie/Bio-Domain-Transfer}.}.
\end{abstract}

\section{Introduction}

Named entity recognition is a crucial step in IE tasks. Existing models have achieved remarkable performance in the general domain~\cite{lin-etal-2020-joint,wang-etal-2021-automated,zhang2023optimizing,shen-etal-2023-promptner}.
However, in the scientific domains, e.g., medical or chemical domains, these models usually struggle due to the extremely large quantity of concepts, the wide presence of multi-token entities, and the ambiguity in detecting entity boundaries.

\begin{figure}
    \centering
    \includegraphics[width=1.0\linewidth]{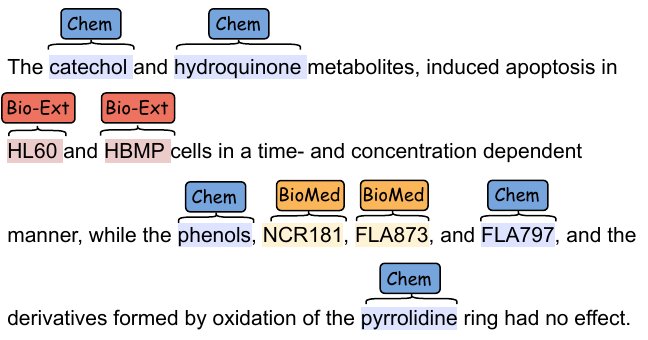}
    \caption{A test example in the chemical domain. The words marked with blue indicators are chemical entities, and the words marked with red and orange indicators are biomedical entities. The entities in red are mislabeled by a few-shot model as chemical entities.} 
    \label{fig:intro_example}
\end{figure}

Large language models (LLMs) show an impressive performance on various NLP tasks such as question answering or text summarization \cite{openai2022chatgpt}. Models such as ChatGPT \cite{openai2022chatgpt} can achieve outstanding results given just a few training examples~\cite{wang-etal-2022-super}. However, \citet{long-tail} recently report that the performance of these models is proportional to the number of relevant documents present in their pretraining corpus. Thus, one can expect that their performance fluctuates across domains. This is particularly expected to occur across certain scientific subjects, where the data may be scarce. Given the already existing challenges of the named entity recognition task in the scientific domain---mentioned earlier---this factor can further exacerbate the problem. For instance, in our early few-shot learning experiments, we observed that the results of ChatGPT in the chemical domain named entity recognition task are significantly worse than those in the general domain.\footnote{We report this complementary experiment in Appendix~\ref{appendix:chatgpt}.} 

In the present work, we employ transfer learning \citep{transfer-survey} to alleviate this problem. Transfer learning methods exploit the label data from one domain (called the source domain) to minimize the prediction error in another domain (namely the target domain). We particularly focus on a realistic setting, where given a large set of labeled data from the source domain and a small set of labeled data from the target domain, the goal is to develop a model for the target domain\footnote{In the literature, this setup is often called the semi-supervised transfer learning setting \citep{semi-super-da}.}. There are more resources in the biomedical domain than in the chemical domain due to funding priorities and BioNLP workshops. Therefore, as a case study, we take the biomedical domain as the source and the chemical domain as the target. 

Figure~\ref{fig:intro_example} shows the challenges a named entity recognition model can face in the chemical domain. The model is trained in a few-shot learning setting. Thus, it is trained on labeled biomedical data, and then, further finetuned on a few labeled examples from the chemical domain. The task is to recognize only the chemical entities---ignoring other types of entities. 
Blue indicators represent chemical entities, while red and orange indicators denote biomedical entities.
The model categorizes the entities marked with blue and red as chemical entities. The first observation is that, to an inexpert human, it is difficult to perform the task because the entities are highly domain-specific. The second observation is that the model wrongfully labels some entities from the source domain as the target entities. 
Therefore, engineers developing such a model face a dilemma: while not using the source data dramatically deteriorates performance,\footnote{We empirically support this in the analysis section.} simply pretraining with the source data increases the false positive rate.
The third, and perhaps the most important, observation is that the examples from the source and target domains can co-occur in the same document. This problem setting contradicts the regular transfer learning setting, where the examples from the source and the target domains are fully disjoint \citep{dom-ada-theory}. The latter property poses difficulties in the applicability of the traditional transfer learning methods in this setting.

Our core idea is to train a named entity recognition model such that it is able to project the representations of the source and target entities into separate regions of the feature space. 
Such a model can potentially transfer knowledge from the source domain to the target domain by constructing a shared feature space between the two domains.
Furthermore, it reduces the similarity between the representations of the entities in the two domains, and consequently, it can potentially minimize the number of source entities that are mislabeled as target entities. To achieve this, our model consists of a pretraining stage on the source data, and a finetuning stage on the target data. 

In the pretraining stage, we propose two methods to enrich the feature space with auxiliary data. The auxiliary data is extracted from the event mentions that the entities participate in. Additionally, during this stage, we propose to employ the multi-similarity loss term \citep{wang2019multi}, which enables us to partition (or group) the source entities. Our empirical analysis shows that constructing such a feature space during the pretraining stage facilitates our projection step during the finetuning stage. In the finetuning stage, we detect the potentially false positive entities by pseudo-labeling them. Then, we aim to construct a feature space that projects the pseudo-labeled entities and the target entities into separate regions. We achieve this by employing the multi-similarity loss again. 
We evaluate our method across twelve use cases and show it outperforms the baselines in most experiments, with improvements of up to 5\% in absolute value.
We also empirically analyze our method and show that each proposed technique is individually effective. Our contributions are threefold:

\begin{itemize}

    \item We propose a new pretraining algorithm for the named entity recognition task in the transfer learning setting. Our algorithm involves two steps: first, extracting auxiliary information about the source entities through the event mentions they participate in; and second, proposing an entity grouping technique using the multi-similarity loss. Our methods have proven effective for the named entity recognition task in the target domain. Our study is carried out in the scientific domain, particularly from the biomedical data as the source domain to the chemical data as the target domain. This is a crucial and challenging real-world problem. All of our claims, here and later, are only about this task.

    \item We propose a finetuning algorithm, which aims to project the target entities and the entities that may be potentially mislabeled into separate regions of the feature space. It comprises two steps: first, detecting the potentially out-of-domain entities by pseudo-labeling them; and second, obtaining dissimilar representations for the two sets of entities using the multi-similarity loss.

    \item We conduct extensive experiments across twelve cases, showing that our method significantly outperforms the baselines and shedding light on various aspects of our model.
\end{itemize}

\begin{figure*}[!htb]
    \centering
    \includegraphics[width=1.0\textwidth]{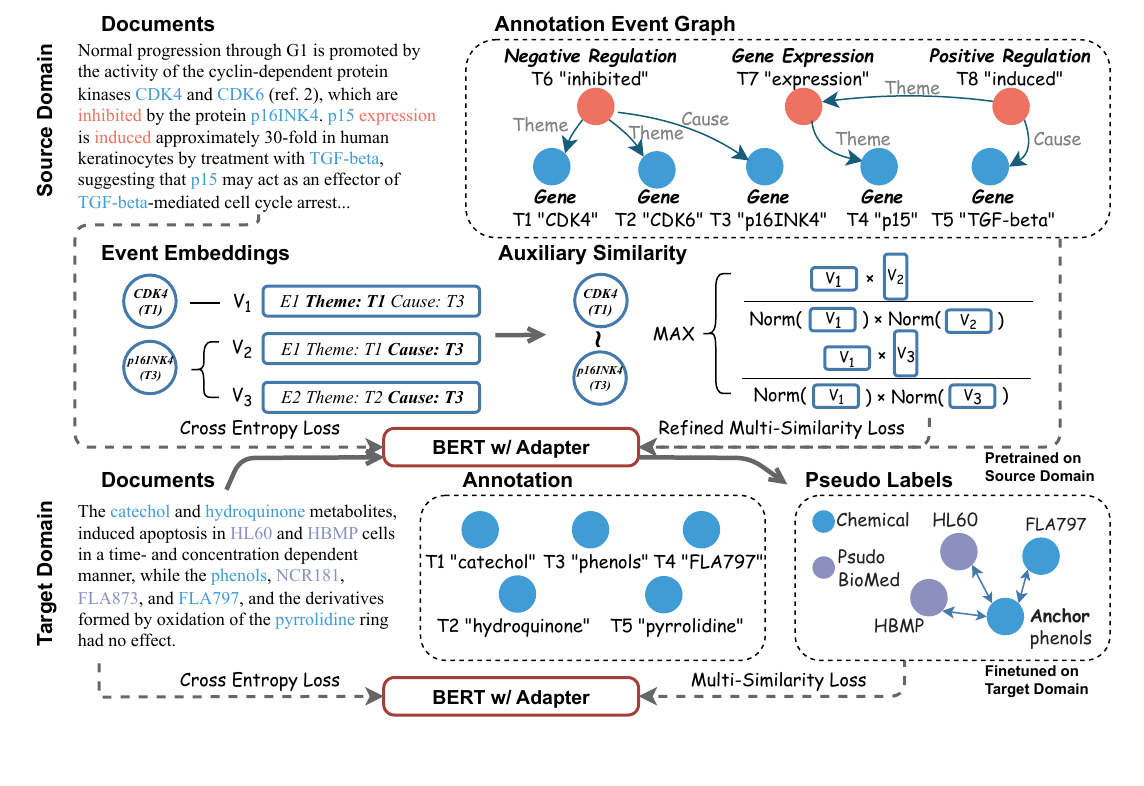}
    \caption{Overview of proposed entity grouping and entity discrimination frameworks. Entity grouping on the source domain is shown in the upper part. Based on event annotations, a set of event embeddings is constructed under two paradigms. Afterward, pairwise auxiliary similarity scores are calculated according to argument embeddings. 
    Extended multi-similarity loss concerning four types of similarities, combined with cross-entropy loss, are jointly learned during pretraining. Entity discrimination on the target domain is shown in the lower part. Pseudo labels are formed by the named entity recognition model pretrained in the source domain, and in contrast to annotated labels, a multi-similarity loss is injected into finetuning.} 
    \label{fig:overview}
\end{figure*}

\section{Background}

\subsection{Named Entity Recognition}

We view the named entity recognition task as a sequence labeling problem. We denote the training data by $D{=}\left\{(\mathbf{X}_i,\mathbf{Y}_i)\right\}_{i=1}^{n}$, where $n$ is the number of input passages (or texts).
To train a  classifier $f$ with parameter $\theta$, we minimize the loss as follows:
\begin{equation}
    \small
    \mathcal{L}_{NER}=\mathbb{E}_{(\mathbf{X}_i,\mathbf{Y}_i) {\sim} D}\left[ CE\left(f(\mathbf{Y}_i|\mathbf{X}_i;\theta), \mathbf{Y}_i\right) \right],
\end{equation}
where $CE$ is the cross-entropy loss.

\subsection{Transfer Learning}

We are given examples from the source domain $\mathbb{S}$ and the target domain $\mathbb{T}$, where the training set size in the target domain is much smaller than the source domain, i.e., $|\mathbf{X}^\mathbb{T}|<<|\mathbf{X}^\mathbb{S}|$. Given this data, we aim to develop a model for the target domain and minimize its prediction error. 

We focus on the named entity recognition task, and take the biomedical domain as the source and the chemical domain as the target. Note that the data distributions in the two domains are different. Therefore, a model solely trained on the source data is usually not as competitive as one trained on the target data. The baseline solution in this setting, \textit{direct transfer}, is to pretrain the model on the labeled source data and finetune it on the labeled target data.

\subsection{Multi-Similarity Loss} 
\label{sub-sec:multi-sim-definition}

We enhance our named entity recognition model in the source domain by capturing the similarities between entity pairs. 
To achieve this, we employ an objective term called the multi-similarity (MS) contrastive loss proposed by \citet{wang2019multi} for metric learning. MS can incorporate self-similarity, relative positive similarity, and relative negative similarity. Self-similarity depends on the properties of the data point itself, such as hardness, while negative and positive similarities are measured with respect to an anchor data point.

In the following sections, we use the final encoder hidden states of input tokens as entity representations. If an entity consists of multiple tokens, we take the average of their representations. Additionally, we denote the relative similarity between entity pairs as $S_{\parm}$, using cosine similarity.

The multi-similarity loss is calculated in two stages. First, given the \textit{i-th} entity denoted by $x_i$ and its label denoted by $y_i$, we aim to extract the most difficult positive and negative entities. This is achieved by thresholding over the relative similarity scores as follows:
\begin{equation}
    \begin{aligned}
        \mathcal{P}_i = \{x_j|S_{ij}^+&<\max_{y_k\neq y_i}S_{ik}+\epsilon\},\\
        \mathcal{N}_i = \{x_j|S_{ij}^-&>\min_{y_k=y_i}S_{ik}-\epsilon\},
    \end{aligned}
\end{equation}
where $\mathcal{P}_i$ is the set of positive, $\mathcal{N}_i$ is the set of negative, and $\epsilon$ is a margin penalty.

Second, we calculate a soft weight score for the extracted pairs to reflect their importance:
\begin{equation}
    \begin{aligned}
        w_{ij}^+&=\frac{e^{-\alpha(S_{ij}-\gamma)}}{1+\sum_{k\in\mathcal{P}_i}e^{-\alpha(S_{ik}-\gamma)}},\\
        w_{ij}^-&=\frac{e^{\beta(S_{ij}-\gamma)}}{1+\sum_{k\in\mathcal{N}_i}e^{\beta(S_{ik}-\gamma)}},
    \end{aligned}
    \label{eq:pair_weights}
\end{equation}
where $\alpha$, $\beta$, and $\gamma$ are hyperparameters. We observe that in each set, the weights are the ratio of the self-similarity scores to the sum of all the relative scores in the set.

The final multi-similarity loss is:
\begin{equation}
\label{eq:ms}
    \begin{aligned}
        \mathcal{L}_{MS}=&\frac{1}{n_e}\sum_{i}^{n_e}\left\{\frac{1}{\alpha}\log[1+\sum_{k\in\mathcal{P}_i}e^{-\alpha(S_{ik}-\gamma)}]\right. \\
        &+\left.\frac{1}{\beta}\log[1+\sum_{k\in\mathcal{N}_i}e^{\beta(S_{ik}-\gamma)}]\right\},
    \end{aligned}
\end{equation}
where $n_e$ is the total number of the entities. Note that the values of $S_{\parm}$ already contain the weights computed in Equation \ref{eq:pair_weights}.

\section{Proposed Method}

Figure \ref{fig:overview} illustrates our framework. It consists of two stages, pretraining on the source data, and then, finetuning on the target data. In the source domain, we employ external knowledge to construct an event feature space based on their arguments, 
and to calculate the auxiliary similarity scores between entities. Then, the similarity scores are used in the multi-similarity loss to shape the entity feature space. In the target domain, we aim to enhance the model's ability to distinguish the target entities from the entities that are likely to be mislabeled, such as the entities that potentially belong to the source domain. To achieve this, we propose an algorithm to extract pseudo-labels and employ the multi-similarity loss for the second time.

\subsection{Source Domain Pretraining} 
\label{sub-sec:source-train}

Using external knowledge, we enrich entity representations for source domain pretraining. Since the source domain consists of a set of various sub-domains (or sub-topics), discovering these sub-domains in the source domain facilitates the subsequent process of domain transfer~\cite{hoffman2012discovering}. Specifically, by detecting these sub-domains and grouping the data, we enable the contrastive loss in the next step to consider each one individually and transform them accordingly. This approach avoids the oversimplification of treating the entire source domain as a single cluster.  Below, we propose two separate approaches to obtain the auxiliary embedding vectors, both exploiting event mentions that the entities appear in. Given the auxiliary vectors, we describe our method for calculating the auxiliary similarity scores. Finally, we provide an overview of the pretraining loss function, which incorporates the entity similarity scores and the auxiliary similarity scores.

\paragraph{Concatenation-based Event Embedding.} 
\label{approach:concat}

Our first approach relies on an off-the-shelf token encoder pretrained on biomedical data, called SapBERT \cite{liu-etal-2021-self}. Given an event mention, we encode its arguments using SapBERT, and then concatenate the resulting vectors to obtain the event representation. Note that in some cases, an argument may be a nested event, or an event may have a varying number of arguments. In those cases, we use vector averaging to compress the representations or padding to fill in the extra argument slots\footnote{The details can be found in Appendix \ref{appendix:concat}.}. 

\paragraph{Sentence-Encoder based Event Embedding.}
\label{approach:sentEnc}
In our second approach, we use templates generated by a LLM. We begin by extracting all event types from the source domain. We then submit each type and its arguments to the LLM, using prompts to construct a template. A few examples of such templates are reported in Table~\ref{tab:sentEnc}, and a larger set of templates along our prompt instruction can be found in Appendix~\ref{appendix:sentEnc}. Then, for the event mentions in the source data, we complete their corresponding templates by replacing the placeholders with the actual arguments. The resulting passages are sent through an off-the-shelf sentence encoder to obtain the final representation vectors. In our experiments, we use ChatGPT \cite{openai2022chatgpt} as the LLM, and use the S-PubMed-BERT~\cite{deka2022improved} as the sentence encoder. 
In \S~\ref{sec:exp}, we individually evaluate each one of the methods for extracting the auxiliary embedding vectors.

\begin{table*}[ht]
    \centering
    \small
    \begin{tabularx}{\linewidth}{>{\centering\arraybackslash\hsize=.3\hsize}X>{\arraybackslash\hsize=1.7\hsize}X}
    \toprule
    \textbf{Type} & \textbf{Template} \\
    \midrule
    Phosphorylation & Indicated by the given trigger <Trigger>, a specific molecule <Theme> is modified by the addition of a phosphate group at a particular site <Site>, facilitated by another molecule <Cause>. \\
    \midrule
    Acetylation & Indicated by the given trigger <Trigger>, a specific molecule <Theme> undergoes the addition of an acetyl group at a particular site <Site>, catalyzed by another molecule <Cause>. \\
    \midrule
    Pathway & Indicated by the given trigger <Trigger>, involving one or more molecules <Participant> that collaborate to accomplish a specific biological function or response. \\
    \bottomrule
    \end{tabularx}
    \caption{Examples of templates for sentence-encoder based embeddings. Angle brackets <·> in templates are placeholders to be replaced by actual entities as corresponding arguments. }
    \label{tab:sentEnc}
\end{table*}

\paragraph{Auxiliary Similarity.}

Given $E(x_i)$ and $E(x_j)$ as the sets of auxiliary vectors for the entities $x_i$ and $x_j$ respectively, we define the similarity between the two entities as the maximum inter-similarity between all the vector pairs across the two sets. More specifically, we formulate it as follows:
\begin{equation}
    \kappa_{ij}=\max_{\mu\in E(x_i),\nu\in E(x_j)} \frac{\mu^T\cdot \nu}{|\mu|\cdot|\nu|}.
\end{equation}
The value of $\kappa_{ij}$ captures the relatedness between the contexts that the two entities appear in. If $E(x_{\parm})$ is empty, then we set $\kappa_{i\parm}{=}0$.

\paragraph{Contrastive Grouping.}

We adapt the multi-similarity loss \cite{wang2019multi} to consider the primary similarity scores between entities, which is the cosine similarity between the encoder outputs for the entities, as well as the auxiliary similarity scores described earlier in this section. Our core idea is to assign a higher weight to the more informative pairs. In the case of positive pairs, this translates into assigning a higher weight to the instances that have a smaller primary similarity and a higher auxiliary similarity. In the case of negative pairs, it is the reverse, i.e., assigning a higher weight to the pairs with higher primary similarity and a lower auxiliary similarity. 

The intuition behind these design choices is as follows. In the case of positive pairs, a low primary similarity and a high auxiliary similarity potentially mean that the encoder is unable to properly project the entities, but there is a strong external signal that the pair must have similar representations. In the case of negative pairs, a high primary similarity and a low auxiliary similarity potentially mean that the model needs to revise the parameters to take into account the external signal.

To implement these ideas, we exploit the soft weights discussed in Equations \ref{eq:pair_weights} to derive the weights for the positive pairs as follows:
\begin{equation}
    \begin{aligned}
        \hat{w}_{ij}^+&=\frac{1}{e^{-I_{ij}^+}+\sum_{k\in\mathcal{P}_i}e^{-J_{ik}^++J_{ij}^+}},\\
        I_{ij}^+&=\alpha(\gamma-S_{ij})+\rho\kappa_{ij},\\ 
        J_{ij}^+&=\alpha S_{ij}-\rho\kappa_{ij}.\\
    \end{aligned}
\end{equation}
The value of $\hat{w}_{ij}^+$ is the second formulation of the soft weights introduced by \citet{wang2019multi}. We see that instead of only relying on the values of $S_{\parm}$ to define $I_{\parm}^+$, we incorporate the auxiliary similarity scores $\kappa_{\parm}$ via the hyperparameter $\rho$.

Similarly, we re-define the soft weights for the negative pairs as follows:
\begin{equation}
    \begin{aligned}
        \hat{w}_{ij}^-&=\frac{1}{e^{I_{ij}^-}+\sum_{k\in\mathcal{N}_i}e^{J_{ik}^--J_{ij}^-}},\\
        I_{ij}^-&=\beta(\gamma-S_{ij})+\tau\kappa_{ij},\\
        J_{ij}^-&=\beta S_{ij}-\tau\kappa_{ij},\\
    \end{aligned}
\end{equation}
where $\tau$ is a hyperparameter to balance the contributions of $S_{\parm}$ and $\kappa_{\parm}$.

Given the re-defined soft weights, the refined multi-similarity objective (RMS) can be re-written as follows:
\begin{equation}
    \begin{aligned}
        \mathcal{L}_{RMS}=&\frac{1}{n_e}\sum_{i}^{n_e}\left\{\frac{1}{\alpha}\log[1+\sum_{k\in\mathcal{P}_i}e^{-\alpha(S_{ik}-\gamma)+\rho\kappa_{ik}}]\right. \\
        &+\left.\frac{1}{\beta}\log[1+\sum_{k\in\mathcal{N}_i}e^{\beta(S_{ik}-\gamma)-\tau\kappa_{ik}}]\right\},
    \end{aligned}
\end{equation}
where, as mentioned earlier, $\rho$ and $\tau$ are to maintain a balance between the primary and the auxiliary representations. Note that we use the information extracted from events to construct the auxiliary representations. However, additional sources of information can be considered if it is present.

The pretraining objective function consists of the supervised named entity recognition term and the unsupervised RMS term, as follows:
\begin{equation}
    \mathcal{L}=\mathcal{L}_{NER}+\lambda_{\mathbb{S}}\mathcal{L}_{RMS},
\end{equation}
where 
$\lambda_{\mathbb{S}}$ is a penalty term.

\subsection{Target Domain Finetuning} 
\label{sub-sec:target-train}

The source and target domains in our problem setting share the same context. In the same documents (or even sentences) that the entities from one domain appear, the entities from the other domain may be used, too. 
This makes the recognition task particularly challenging in the target domain for two reasons: the training data in this domain is scarce, and the presence of entities from the source domain can potentially lead to a high false positive rate.
Our core idea is that, while finetuning the model on the target data, we train the encoder such that it projects the target entities and the entities that potentially belong to the source domain into separate regions of the feature space. To implement this idea, we employ pseudo labeling along the multi-similarity loss--introduced in \S \ref{sub-sec:multi-sim-definition}.

\paragraph{Pseudo Labeling.}

Given a passage with annotated target entities in the target training data, we use the model introduced in \S \ref{sub-sec:source-train} to automatically detect the entities that may belong to the source domain. These entities act as pseudo labels in our algorithm. Note that while there may be passages that do not contain such entities, in general, due to the nature of the two domains that we are studying (i.e., biomedical and chemical domains), this is an expected observation. In the results section, we will also empirically support our argument.

\paragraph{Contrastive Discrimination.}

In the next step, we enable the model to discriminate between the target and pseudo-labeled source entities. For this purpose, we use the multi-similarity loss. 
We use multi-similarity loss in Eq.~\ref{eq:ms} to calculate contrastive objective by defining the labels as follows:
\begin{equation}
    y_i=\left\{
         \begin{array}{lr}
         0, & x_i\in \mathcal{Q} \\  
         1, & x_i\notin \mathcal{Q}
         \end{array}
    \right.
\end{equation}
where $x_i$ represents the entity and $\mathcal{Q}$ is the set of entities with pseudo labels.

The final target domain fine-tuning objective is:
\begin{equation}
    \mathcal{L}=\mathcal{L}_{NER}+\lambda_{\mathbb{T}}\mathcal{L}_{MS},
\end{equation}
where $\lambda_{\mathbb{T}}$ is a penalty terms.

\section{Experiments}
\label{sec:exp}

\subsection{Experimental Setup}

\begingroup
\setlength{\tabcolsep}{0pt} 
\begin{table}[!htb]
    \centering
    \small
    \begin{tabularx}{\linewidth}{>{\centering\arraybackslash\hsize=0.6\hsize}X>{\centering\arraybackslash\hsize=1.15\hsize}X>{\centering\arraybackslash\hsize=1.25\hsize}X}
    \toprule
    \textbf{Dataset} & \# Train/ Valid/ Test & \# Train/ Valid/ Test (All) \\
    \toprule
    PC & -  & 260 / 89 / 175 \\
    ID & -  & 150 / 46 / 117 \\
    CG & -  & 300 / 100 / 200 \\
    \midrule
    CHEMDNER  & 81 / 72 / 2478 & 2916 / 2907 / 2478 \\
    BC5CDR  & 86 / 88 / 500 & 500 / 500 / 500 \\
    DrugProt  & 93 / 88 / 600 & 597 / 597 / 600 \\
    \bottomrule
    \end{tabularx}
    \caption{Overview of the datasets. The top three are the biomedical source datasets, and the bottom three are the chemical target datasets. The target datasets were down-sampled randomly to be used in the few-shot setting.}
    \label{tab:datasets}
\end{table}
\endgroup

\begingroup
\begin{table*}[!htb]
    \centering
    \footnotesize
    \begin{tabularx}{\linewidth}{>{\centering\arraybackslash\hsize=1.72\hsize}X|>{\centering\arraybackslash\hsize=.92\hsize}X>{\centering\arraybackslash\hsize=.92\hsize}X>{\centering\arraybackslash\hsize=.92\hsize}X>{\centering\arraybackslash\hsize=.92\hsize}X>{\centering\arraybackslash\hsize=.92\hsize}X>{\centering\arraybackslash\hsize=.92\hsize}X>{\centering\arraybackslash\hsize=.92\hsize}X>{\centering\arraybackslash\hsize=.92\hsize}X>{\centering\arraybackslash\hsize=.92\hsize}X}
    \toprule
      \textbf{Target Tasks} & \multicolumn{3}{c}{\textbf{\textit{CHEMDNER}}} & \multicolumn{3}{c}{\textbf{\textit{BC5CDR}}} & \multicolumn{3}{c}{\textbf{\textit{DrugProt}}} \\
      \textbf{Metrics} & \textbf{Precision} & \textbf{Recall} & \textbf{F1} & \textbf{Precision} & \textbf{Recall} & \textbf{F1} & \textbf{Precision} & \textbf{Recall} & \textbf{F1} \\
    \midrule
    Target Only & 42.73 & 51.69 & 46.77 & 72.44 & 85.86 & 78.51 & 63.80 & 67.42 & 65.49 \\
    Direct Transfer & 44.11 & 48.60 & 46.18 & 71.92 & 86.54 & 78.55 & 66.36 & 70.13 & 68.17 \\
    \midrule
    EG(concat) & 43.63$\pm$1.7 & 51.06$\pm$0.1 & 47.03$\pm$1.0 & 72.10$\pm$1.3 & 87.12$\pm$0.5 & 78.89$\pm$0.6 & 68.49$\pm$0.9 & 66.54$\pm$0.9 & 67.49$\pm$0.2 \\
    EG(sentEnc) & 43.09$\pm$3.7 & 51.60$\pm$3.3 & 46.91$\pm$2.9 & 73.96$\pm$1.1 & 86.56$\pm$0.8 & 79.76$\pm$0.3 & 66.02$\pm$0.5 & 69.97$\pm$0.8 & 67.93$\pm$0.3 \\
    ED & \textbf{45.71}$\pm$2.3 & 50.13$\pm$1.2 & 47.78$\pm$1.3 & 75.01$\pm$2.0 & \textbf{87.80}$\pm$0.5 & 80.88$\pm$0.9 & \textbf{70.15}$\pm$0.3 & \textbf{71.79}$\pm$0.3 & \textbf{70.96}$\pm$0.1 \\
    EG(concat)+ED & 43.83$\pm$0.6 & \textbf{52.17}$\pm$0.9 & 47.64$\pm$0.7 & 73.70$\pm$0.8 & 85.50$\pm$0.9 & 79.15$\pm$0.1 & 68.01$\pm$0.2 & 69.06$\pm$1.3 & 68.52$\pm$0.6 \\
    EG(sentEnc)+ED & 45.28$\pm$0.4 & 51.68$\pm$1.8 & \textbf{48.26}$\pm$0.9 & \textbf{76.06}$\pm$1.8 & 86.58$\pm$1.3 & \textbf{80.95}$\pm$0.5 & 67.16$\pm$0.8 & 69.04$\pm$1.0 & 68.08$\pm$0.9 \\
    \bottomrule
    \end{tabularx}
    \caption{Evaluation results precision, recall, and F1(\%) scores on three target tasks with Biomedical Multi-task as source task. All the reported scores are averaged over 3 different random seeds. We include two baselines, along with our methods EG (Entity Grouping), ED (Entity Discrimination), and their combination.}
    \label{tab:main}
\end{table*}
\endgroup

\paragraph{Datasets.} As the source datasets, we use three benchmarks: Pathway Curation (PC), Cancer Genetics (CG), and Infectious Diseases (ID). The first two datasets were released by BioNLP Shared Task 2013 \cite{nedellec-etal-2013-overview}, and the third one was released by BioNLP Shared Task 2011 \cite{pyysalo-etal-2011-overview}. All three datasets have the same format. We aggregate them to create a fourth dataset called the \textit{biomedical multi-task} dataset. As the target datasets, we use three benchmarks: CHEMDNER \cite{krallinger2015chemdner}, BC5CDR \cite{kim2015overview}, and DrugPort \cite{miranda2021overview}.

Each dataset contains extra annotations unrelated to its domain.
For instance, the CG dataset has annotations not relevant to biomedicine. We pre-process all the datasets by removing such annotations. For few-shot experiments, we down-sample the training and validation sets of target datasets to sizes randomly chosen between 70 and 100.
Table \ref{tab:datasets} summarizes the dataset statistics.

\paragraph{Baselines.} We compare our method with two baselines. \textit{Target Only}: a model finetuned on the labeled target data. \textit{Direct Transfer}: a model pretrained on the labeled source data and then finetuned on the labeled target data.

\noindent\paragraph{Implementation Details.} We use BERT (bert-base-uncased) \cite{devlin-etal-2019-bert} as the backbone for all the models. To train the model, we update the parameters of the adapter layers \cite{houlsby2019parameterefficient} and freeze the rest, due to limited computational resources.
We iteratively select each source and target dataset pair as the training and evaluation benchmarks. 
All the experiments are repeated three times. 
Following \citet{seqeval}, we report average macro Precision, Recall, and F1 scores\footnote{Training and tuning details can be found in Appendix~\ref{appendix:exp_detail}.}.

\subsection{Main Results}

Table \ref{tab:main} reports the performance of our model compared to the \textit{target Only} and the \textit{Direct Transfer} models, when the dataset \textit{biomedical multi-task} is used as the source data. All the other use cases are reported in Appendix \ref{appendix:full_res}. Our final model EG($\parm$)+ED outperforms the baselines in the majority of the cases. We also report the performance of each component of our method in the table---i.e., our entity grouping (EG) and entity discrimination (ED) techniques. We observe that, on average, the method further improves when they are both used in the pipeline. One exception is the DrugProt dataset, which we discuss in the next section.

\begingroup
\begin{table}
    \centering
    \small
    \begin{tabularx}{\linewidth}{>{\centering\arraybackslash\hsize=1.3\hsize}X>{\centering\arraybackslash\hsize=.9\hsize}X>{\centering\arraybackslash\hsize=.9\hsize}X>{\centering\arraybackslash\hsize=.9\hsize}X}
    \toprule
    \textbf{Target Tasks} & \textbf{\textit{CHEMD}} & \textbf{\textit{BC5CDR}} & \textbf{\textit{DrugProt}} \\
    \toprule
    Direct Transfer & 46.18 & 78.55 & 68.17 \\
    \midrule
    Pse-Augment & 47.13$\pm$0.8 & 76.43$\pm$0.2 & 68.20$\pm$0.8 \\
    Pse-Classifier & 46.93$\pm$1.4 & 78.80$\pm$0.3 & 69.42$\pm$0.1 \\
    Ours & \textbf{47.78}$\pm$1.3 & \textbf{80.88}$\pm$0.9 & \textbf{70.96}$\pm$0.1 \\
    \bottomrule
    \end{tabularx}
    \caption{F1(\%) scores of our method compared to two alternative methods for using pseudo-labels. All the reported scores are averaged over 3 different random seeds. Additional experiment details are in Appendix~\ref{appendix:ablation_exp_detail}.}
    \label{tab:ed_ablation}
\end{table}
\endgroup

\subsection{Empirical Analysis}

\paragraph{Pseudo Label Usage.}
Table \ref{tab:ed_ablation} reports a comparison between our model and two alternative methods. \textit{Pse-Augment}, where the detected pseudo-labels are marked and augmented with the target entities and a classifier trained to label unseen target entities. \textit{Pse-Classifier}, where a classifier is trained to detect pseudo-labels and to filter them out, before being potentially mislabeled. This experiment aims to reveal the efficacy of the multi-similarity (MS) loss for discriminating between the source and the pseudo-labeled entities. The results indicate that our ED method that leverages MS loss is an effective way to use pseudo labels. 
For \textit{Pse-Augment}, adding source domain entity labels to the target task leads to the negative transfer (NT) problem~\cite{Zhang2020OvercomingNT}. For \textit{Pse-Classifier}, the pipeline suffers from error propagation, where errors caused by the classifier can severely affect the performance of target entity predictions.

\begin{figure}
    \centering
    \includegraphics[width=\linewidth]{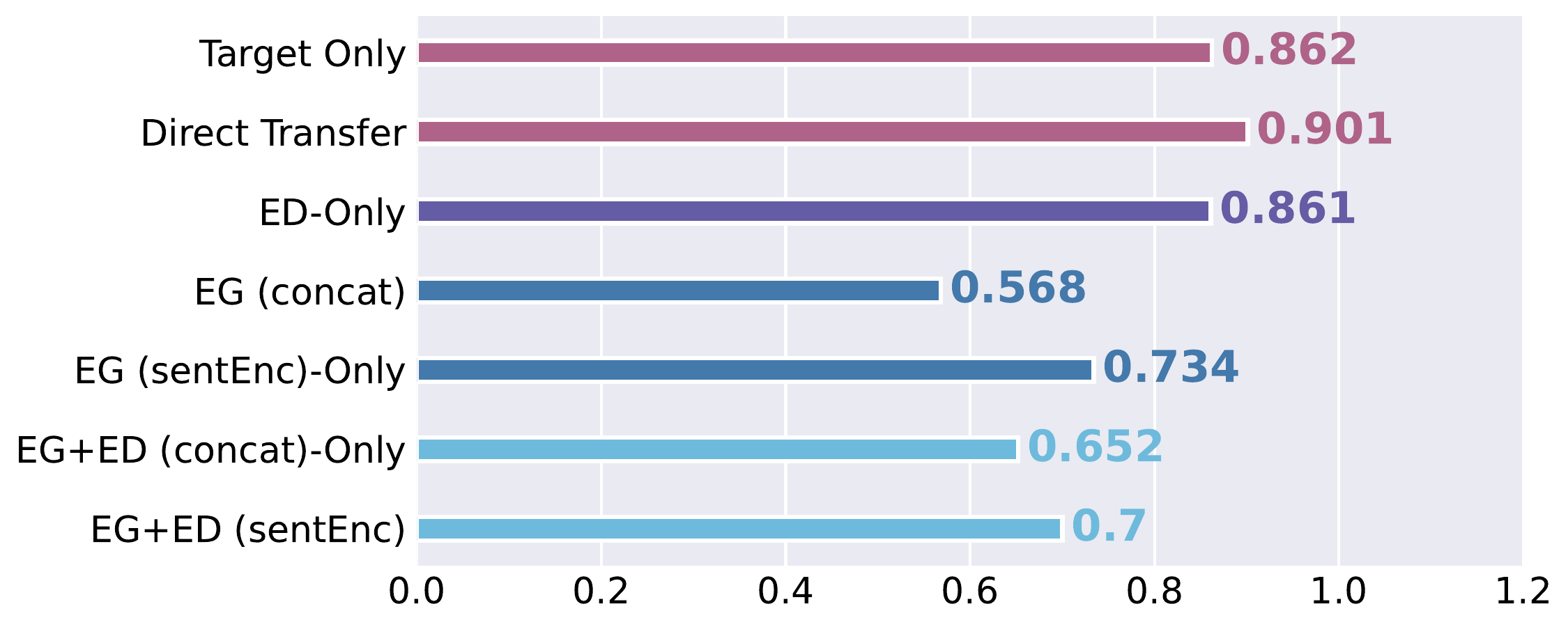}
    \caption{Davies-Bouldin index criterion of clusters. For baseline and ED-concerned settings, pseudo entities are included and viewed in the same cluster as \textit{Disease}.}
    \label{fig:analysis_inertia}
\end{figure}

\paragraph{How is the Representation Enhanced?}
To investigate the effect of our proposed methods, we project entity representations (i.e., averaged hidden states of entity tokens in input texts) into a two-dimensional space using t-SNE \cite{Maaten2008VisualizingDU}\footnote{We show the visualization of target task BC5CDR in Appendix \ref{appendix:visual}.}. For better comparison, we adopt the Davies-Bouldin index (DB) \cite{cluster_measure} as the criterion. The lower the DB, the better the clustering of the data points.

With our ED method, the model effectively learns to disperse the representations of chemical entities and pseudo-labeled entities. Therefore, it becomes easier for our model to assign a negative label to a source domain entity by measuring similarities between its representations and the target domain entity representations. Furthermore, our EG method also plays a vital role in the formation of feature space of the target domain. The projection results in clearer clustering of chemical entities, while disease entities are dispersed from them, making it easier to extract chemical entities. For a more precise comparison, Figure \ref{fig:analysis_inertia} reports the DB index. Our methods achieve lower DB compared to the baselines, which indicates that the improved representations of the target domain are learned with our methods.

\begin{table}[htb]
    \centering
    \small
    \begin{tabularx}{\linewidth}{>{\centering\arraybackslash\hsize=1.3\hsize}X>{\centering\arraybackslash\hsize=0.9\hsize}X>{\centering\arraybackslash\hsize=0.9\hsize}X>{\centering\arraybackslash\hsize=0.9\hsize}X}
    \toprule
    \textbf{Method} & \textbf{Precision} & \textbf{Recall} & \textbf{F1} \\
    \toprule
    Target Only   &  41.32$\pm$1.9 & 45.52$\pm$2.0 & 43.31$\pm$1.9 \\
    \toprule
    Direct Transfer & 42.16$\pm$1.3 & 48.51$\pm$1.4 & 45.08$\pm$0.6 \\
    EG(sentEnc)      & 45.49$\pm$1.0 & 49.03$\pm$1.9 & 47.16$\pm$0.8 \\
    ED               & 45.35$\pm$3.2 & 47.88$\pm$3.5 & 46.58$\pm$3.3 \\
    EG(sentEnc)+ED   & 44.94$\pm$0.4 & 49.37$\pm$2.0 & 47.03$\pm$0.8 \\
    \bottomrule
    \end{tabularx}
    \caption{Averaged F1 scores (\%) over 3 different random seeds for CHEMDNER trained on full BERT model.}
    \label{tab:full_model}
\end{table}

\paragraph{Role of Adapters}
To clarify that the use of adapters does not interfere with the conclusion of our proposed methods, we additionally finetune the full BERT model on the CHEMDNER dataset, and the results are reported in Table~\ref{tab:full_model}. The performance of the full model is similar to the performance of adapters or even slightly worse than our adapter models (47.03 vs 48.26). Our methods remain effective when fine-tuned with the full model, demonstrating that the results in the paper are reliable and sufficient.

\begin{table}[htb]
    \centering
    \small
    \begin{tabularx}{\linewidth}{>{\centering\arraybackslash\hsize=0.6\hsize}X>{\centering\arraybackslash\hsize=1\hsize}X>{\centering\arraybackslash\hsize=1.4\hsize}X}
    \toprule
    \textbf{Domain} & \#Train/Valid/Test & \#Train/Valid/Test (All) \\
    \toprule
    Science & 28 / 66 / 543 & 200 / 450/ 543 \\
    AI & 13 / 46 / 431 & 100 / 350 / 431 \\
    \bottomrule
    \end{tabularx}
    \caption{Overview of CrossNER dataset.}
    \label{tab:crossner_dataset}
\end{table}
\begin{table}[htb]
    \centering
    \small
    \begin{tabularx}{\linewidth}{>{\centering\arraybackslash\hsize=1\hsize}X>{\centering\arraybackslash\hsize=1\hsize}X>{\centering\arraybackslash\hsize=1\hsize}X>{\centering\arraybackslash\hsize=1\hsize}X}
    \toprule
    \textbf{Target Only} & \textbf{Direct Transfer} & \textbf{EG(sentEnc)} & \textbf{ED} \\
    \toprule
    0.59 & 16.74 & 19.12 & 17.18 \\
    \bottomrule
    \end{tabularx}
    \caption{Averaged F1 scores (\%) over three different random seeds for the CrossNER dataset, transferring from the Science domain to the AI domain.}
    \label{tab:crossner_res}
\end{table}

\paragraph{Compatibility Across Other Domain Pairs}
In the above experiments, we focus on transfer learning between the biomedical domain and the chemical domain. To show the generalization ability of our proposed framework on other domain pairs, we conduct the experiments on two additional domains based on CrossNER~\cite{liu2021crossner}, transferring from the Science domain to the AI domain. These two domains are highly related and share similar named entities. We downsample the train and validation data to roughly 10\% of the full dataset for the target domain. Detailed statistics are shown in Table~\ref{tab:crossner_dataset}. The F1 scores averaged over three runs are reported in Table~\ref{tab:crossner_res}. It shows that our methods have strong generalization ability.

\section{Related Work}

\paragraph{Biomedical and Chemical Entity Extraction.}
Entity extraction is a primary step in facilitating scientific discovery~\cite{wang-etal-2021-covid}. Previous biomedical entity extraction methods can be categorized into several classes, including domain adaptive pretraining~\cite{labrak-etal-2023-drbert}, boundary denoising diffusion model~\cite{shen-etal-2023-diffusionner}, question answering-based classification~\cite{arora-park-2023-split}, Cocke-Younger-Kasami (CYK) algorithm~\cite{corro-2023-dynamic}, in-context learning~\cite{chen-etal-2023-learning}, synthetic data~\cite{khandelwal-etal-2022-biomedical,chen-etal-2022-style,hiebel-etal-2023-synthetic}, and prototype learning~\cite{cao-etal-2023-gaussian}. 

Although there's a shared corpus between the biomedical and chemical domains, entity extraction in the chemical domain remains underexplored.
The chemical entity extraction task is usually viewed as an auxiliary task for biomedical named entity recognition~\cite{phan2021scifive,sparknlp,biored,lee-etal-2023-autotrigger}. Similar to \citet{nguyen-etal-2022-hardness} and \citet{wang-etal-2024-chem}, our paper differs from previous papers by viewing the chemical domain as an independent subject. Previous methods try to address this task by distant-supervision~\cite{wang-etal-2021-chemner} or span-representation learning~\cite{nguyen-etal-2023-span}. On the contrary, given the shared corpus between the biomedical and chemical domains, we leverage the large labeled data in the biomedical domain through transfer learning.

\paragraph{Transfer Learning for Named Entity Recognition.}
Transfer learning is an effective method to address low-resource named entity recognition tasks~\cite{lee-etal-2018-transfer,cao-etal-2018-adversarial} and has shown its effectiveness in the biomedical domain~\cite{peng-etal-2019-transfer}.
Prior work has explored the role of continual pretraining on the target domain data~\cite{gururangan-etal-2020-dont,liu2021crossner}. However, \citet{mahapatra-etal-2022-entity} argues that continual pretraining is inefficient regarding computational resources. In contrast to domain-adaptive pretraining, we aim to improve the representation of entities by projecting the source and target entities into separate regions of feature space. Inspired by the success of incorporating external knowledge for biomedical information extraction~\cite{zhang-etal-2021-fine,bioacm}, we use biomedical events to augment the representations. To separate the potentially false positive examples in the target domain, we introduce pseudo-labels. Previous papers have adopted pseudo-labels in cross-lingual named entity recognition~\cite{zhou-etal-2023-improving, ma-etal-2023-colada}. However, they aim to align the entities in the source and target language instead of separating source entities from target entities. Compared to our method, \citet{zhou-etal-2023-improving} has a different contrastive objective, which aims to separate different entity types, rather than the entities from the two domains.

\section{Conclusions}

We proposed a named entity recognition task for transferring knowledge from the biomedical domain to the chemical domain. Our core idea is to train a shared feature space between the two domains to facilitate the knowledge transfer, and, at the same time, to project the source and target entities into separate regions of the feature space to reduce the false negative rate. We achieve this in a few steps. We begin by enriching the source feature space with information about events, then train a named entity recognition model to cluster similar entities into groups. We then use the trained model to label the entities that may belong to the source domain, and use these entities in a multi-similarity loss function to achieve our goal. Our experiments across three sources and three target datasets signify the effectiveness of our method.

\section{Limitations}

Our method partly relies on external knowledge. Therefore, the quality of external knowledge significantly influences the effect of our method. Especially when human annotations are unavailable, the performance of automatic annotators, typically neural networks, is an important factor to consider. 

In this paper, we propose a framework that incorporates external knowledge during training. For instance, the compression function in \S \ref{approach:concat} and templates in Section \ref{appendix:sentEnc} can be altered. Besides, such designs require prior knowledge of the source domain.

Our method leverages a contrastive learning strategy. However, the training algorithm doesn't fully utilize GPU resources, leading to training inefficiencies.

\section*{Acknowledgements}
 This work is supported by U.S. DARPA ITM FA8650-23-C-7316, by the Molecule Maker Lab Institute: an AI research institute program supported by NSF under award No. 2019897, by DOE Center for Advanced Bioenergy and Bioproducts Innovation U.S. Department of Energy, Office of Science, Office of Biological and Environmental Research under Award Number DESC0018420, by U.S. the AI Research Institutes program by National Science Foundation and the Institute of Education Sciences, Department of Education through Award No. 2229873 - AI Institute for Transforming Education for Children with Speech and Language Processing Challenges, and by AI Agriculture: the Agriculture and Food Research Initiative (AFRI) grant no. 2020-67021- 32799/project accession no.1024178 from the USDA National Institute of Food and Agriculture. The views and conclusions contained herein are those of the authors and should not be interpreted as necessarily representing the official policies, either expressed or implied of, the National Science Foundation, the U.S. Department of Energy, and the U.S. Government. The U.S. Government is authorized to reproduce and distribute reprints for governmental purposes notwithstanding any copyright annotation therein.
\bibliography{anthology,custom}

\appendix

\begin{table*}[!htb]
    \centering
    \small
    \begin{tabularx}{\linewidth}{>{\centering\arraybackslash\hsize=0.1\hsize}X>{\arraybackslash\hsize=1.9\hsize}X}
    \toprule
    \textbf{Role} & \textbf{Prompt} \\
    \toprule
    System & You are an expert of chemical named entity recognition tasks\\
    \midrule
    \multirow{11}{*}{User} & Description: In this task, you are given a small paragraph of a PubMed article, and your task is to identify all the named entities (particular chemical related entity) from the given input and also provide type of the each entity according to structure-associated chemical entity mention classes (ABBREVIATION, IDENTIFIER, FORMULA, SYSTEMATIC, MULTIPLE, TRIVIAL, FAMILY). Specifically, the paragraph are given with seperate tokens and you need to list all the chemical named entities in order and also tag their types. Generate the output in this format: entity1 <type\_of\_entity1>, entity2 <type\_of\_entity2>. \\
     & Examples: \\
     & Input: In situ C-C bond cleavage of vicinal diol following by the lactolisation resulted from separated treatment of Arjunolic acid ( 1 ) , 24-hydroxytormentic acid ( 2 ) and 3-O-β-D-glucopyranosylsitosterol ( 3 ) with sodium periodate and silica gel in dried THF according to the strategic position of hydroxyl functions in the molecule . \\
     & Output: C-C <FORMULA>, vicinal diol <FAMILY>, Arjunolic acid <TRIVIAL>, 24-hydroxytormentic acid <SYSTEMATIC>, 3-O-β-D-glucopyranosylsitosterol <SYSTEMATIC>, sodium periodate <SYSTEMATIC>, silica gel <TRIVIAL>, THF <ABBREVIATION>, hydroxyl <SYSTEMATIC> \\
     & Input: Structural studies using LC/MS/MS analysis and ( 1 ) H NMR spectroscopy showed the formation of a glycosidic bond between the primary hydroxyl group of RVX-208 and glucuronic acid . \\
     & Output: ( 1 ) H <FORMULA>, primary hydroxyl <FAMILY>, RVX-208 <IDENTIFIER>, glucuronic acid <TRIVIAL> \\
     & Input: The lystabactins are composed of serine ( Ser ) , asparagine ( Asn ) , two formylated/hydroxylated ornithines ( FOHOrn ) , dihydroxy benzoic acid ( Dhb ) , and a very unusual nonproteinogenic amino acid , 4,8-diamino-3-hydroxyoctanoic acid ( LySta ) . \\
     & Output: lystabactins <FAMILY>, serine <TRIVIAL>, Ser <FORMULA>, asparagine <TRIVIAL>, Asn <FORMULA>, formylated/hydroxylated ornithines <MULTIPLE>, FOHOrn <ABBREVIATION>, dihydroxy benzoic acid <SYSTEMATIC>, Dhb <ABBREVIATION>, 4,8-diamino-3-hydroxyoctanoic acid <SYSTEMATIC>, LySta <ABBREVIATION> \\
     & Please continue: \\
     & Input: \%s \\
     & Output: \\
    \bottomrule
    \end{tabularx}
    \caption{Prompts of ChatGPT evaluation. \%s is to be replaced by test context.}
    \label{tab:chatgpt_config}
\end{table*}
\section{Details of ChatGPT evaluation on CHEMDNER}
\label{appendix:chatgpt}

\begin{figure}[!htb]
    \centering
    \includegraphics[width=0.45\textwidth]{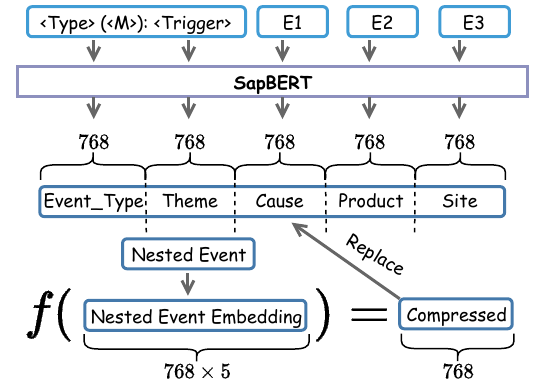}
    \caption{Components of concatenation based event embeddings. Arguments of events, along with event type, are encoded by an off-the-shelf model and concatenated afterwards. For nested events as arguments, we fill in compressed event embeddings recursively.
    \label{fig:concat1}}
\end{figure}

To evaluate ChatGPT's few-shot ability of named entity recognition in the chemical domain, we elaborately select three named entity recognition examples in the CHEMDNER dataset that include all types of entities. An instructional description of the CHEMDNER task and examples constitute the prompt used to instruct the prediction. Configuration of prompts is shown in Table \ref{tab:chatgpt_config}. The precision, recall, and macro-F1 scores on the CHEMDNER test set are $11.09\%$, $35.37\%$, and $16.88\%$ respectively. \citet{laskar-etal-2023-systematic} reports the ChatGPT's named entity recognition performance on general domain named entity recognition task-WNUT 17 dataset \cite{derczynski-etal-2017-results} with precision: $18.03\%$, recall: $56.16\%$, and F1: $27.03\%$. This highlights ChatGPT's weaker performance in solving chemical domain named entity recognition tasks.

\section{Notation Table}

We present definitions for all notations we used in Table \ref{tab:notation}.

\begin{table}[!htb]
    \centering
    \small
    \begin{tabularx}{\linewidth}{>{\centering\arraybackslash\hsize=0.3\hsize}X>{\arraybackslash\hsize=1.7\hsize}X}
    \toprule
    \textbf{Notation} & \textbf{Definition} \\
    \toprule
    $\mathbf{X}$ & Set of input documents \\
    $\mathbf{Y}$ & Set of named entity tags \\
    $n_e$ & Total number of entities \\
    $x_i$ & The $i$-th entity \\
    $y_i$ & Label of the $i$-th entity \\
    $f$ & Classifier model \\
    $\mathbb{S}$ & Source domain \\
    $\mathbb{T}$ & Target domain \\
    $S_{\parm}$ & Relative similarity \\
    $E(\parm)$ & Set of embedding vectors \\
    $\kappa_{\parm}$ & Auxiliary similarity \\
    $\mathcal{P}_i$ & Set of selected positive entities \\
    $\mathcal{N}_i$ & Set of selected negative entities \\
    $\mathcal{Q}$ & Set of entities with pseudo labels \\
    $I^+$, $J^+$ & Auxiliary term for positive pairs \\
    $I^-$, $J^-$ & Auxiliary term for negative pairs \\
    $w^+$ & Soft weight of positive pairs \\
    $w^-$ & Soft weight of negative pairs \\
    $\hat{w}_{ij}^+$ & Adjusted soft weight of positive pairs \\
    $\hat{w}_{ij}^-$ & Adjusted soft weight of negative pairs \\
    $\mathcal{L}_{NER}$ & Classification loss for NER task \\
    $\mathcal{L}_{MS}$ & Multi-similarity loss \\
    $\mathcal{L}_{RMS}$ & Refined multi-similarity loss \\
    $\theta$ & Classifier parameter \\
    $\epsilon$ & Margin penalty \\
    $\alpha$, $\beta$, $\gamma$, $\rho$, $\tau$ & Hyperparameters \\
    $\lambda_{\mathbb{S}}$, $\lambda_{\mathbb{T}}$ & Balance term for losses \\
    \bottomrule
    \end{tabularx}
    \caption{Notation Table}
    \label{tab:notation}
\end{table}

\section{Details of Event Embedding Construction.}

\subsection{Concatenation based Event Embedding}
\label{appendix:concat}

An overview of concatenation based event embedding strategy is shown in Figure~\ref{fig:concat1}. Each type of annotated event contains a trigger and various arguments, including theme, cause, product, and site. We prepare raw texts of each argument and encode them into argument embeddings. To illustrate the formation of raw texts, let's consider an example. Imagine a gene named ``IL-1ra'' which is associated with two event annotations, ``M1, Negation, E9'' and ``E9,	Binding:forms a complex, Theme:IL-1ra, Theme2:Type I IL-1R''. Raw text for ``event\_type'' comprises the event name, ``M'' label, and trigger. The example's raw text should be ``Binding (Negation): forms a complex''. Raw text typically is the corresponding entity itself for the rest of the arguments. However, for the focusing entity, ``IL-1ra'' for instance, raw text is specified as ``self'', deriving ``IL-1ra (self)'' in this case. 
There are several corner cases to tackle with:

\noindent\paragraph{Nested Event.}
As mentioned in \S \ref{approach:concat}, for the nested event, we first recursively compose the event embedding for it and compress it to the same length as partial embedding. To achieve this, let's consider the nested event embedding as $e$. We then implement the compression function by averaging several successive elements:
\begin{equation}
    f(e)=\left[\frac{1}{5}\sum_{k=0}^{4}e_k, \frac{1}{5}\sum_{k=5}^{9}e_k,...,\frac{1}{5}\sum_{k=5i}^{5i+4}e_k,...\right]^T,
\end{equation}
Full embedding has $768\times 5$ dimensions, and we average every $5$ element and concatenate the values into a $768$-dimension embedding. 

\noindent\paragraph{Padding}
It is necessary that some arguments do not apply to some events or miss in annotation. A padding scheme is necessary for missing arguments, and we choose to fill in random partial embedding with the same length sampled from Gaussian distribution with the same mean and covariance as all other encoded partial embeddings.

\subsection {Sentence-Encoder based Event Embedding}
\label{appendix:sentEnc}

The prompt we use to instruct ChatGPT to generate explanatory templates for events is: \textit{give a one-sentence definition of biomedical event type XXX with arguments XXX, XXX...}. For instance, we generate a template for the Phosphorylation event with \textit{give a one-sentence definition of biomedical event type Phosphorylation with arguments Trigger, Theme:Molecule,  Cause:Molecule, Site:Simple chemical}. The full list of the used templates is shown in Table~\ref{tab:full_temp_pc_0}, \ref{tab:full_temp_pc_1}, \ref{tab:full_temp_id_0}, \ref{tab:full_temp_cg_0}, and \ref{tab:full_temp_cg_1}.

\section{Experiment Details}
\label{appendix:exp_detail}

We select \textit{bert-base-uncased} version of BERT model as our backbone model, and we only train 0.817\% of parameters (894,528) of the entire model using transformer-adapter utils~\cite{pfeiffer-etal-2020-adapterhub}. For source task pretraining, we use a batch size of 64; while for target task fine-tuning, we use a batch size of 16, considering the relatively small training set. We use AdamW optimizer and an initial learning rate of $1e-4$ for pretraining and finetuning. To fully train our model, we first train 80 epochs and then stop when f1 scores on the held-out validation set fail to update the best score for at least 20 epochs in a row. For hyperparameters, we tune balance factor $\lambda_\mathbb{S}$ from scale $\{0.10, 0.15, 0.20, 0.25, 0.30\}$ and tune $\lambda_\mathbb{T}$ from scale $\{0.6, 0.8, 1.0, 1.2, 1.4\}$. Due to the limitation of computational resources, we select $\epsilon$ and $\gamma$ as $0.1$ and $0.5$ respectively, and $\alpha$, $\beta$, $\rho$, $\tau$ as $4.0$, $3.0$, $8.0$, $6.0$ respectively considering the ratio of number of positive and negative pairs in contrastive learning.

Our models are trained on 4 Nvidia RTX 2080Ti GPUs in a data parallel fashion. Source task pretraining with contrastive learning takes around 5 hours, while target task finetuning with contrastive learning takes around 30 minutes.

\section{Full Experiment Results}
\label{appendix:full_res}

Full evaluation results are reported in Table \ref{tab:full_res}.

\begingroup
\begin{table}[htb]
    \centering
    \small
    \begin{tabularx}{\linewidth}{>{\centering\arraybackslash\hsize=2.05\hsize}X>{\centering\arraybackslash\hsize=.65\hsize}X>{\centering\arraybackslash\hsize=.65\hsize}X>{\centering\arraybackslash\hsize=.65\hsize}X}
    \toprule
    \textbf{Target Tasks} & \textbf{\textit{CHEMD}} & \textbf{\textit{BC5CDR}} & \textbf{\textit{DrugProt}} \\
    \toprule
    Direct Transfer & 46.18 & 78.55 & \textbf{68.17} \\
    EG(MS) & 46.08 & 78.87 & 68.00 \\
    \midrule
    EG(concat) & \textbf{47.03} & 78.89 & 67.49 \\
    EG(sentEnc) & 46.91 & \textbf{79.76} & 67.93 \\
    \bottomrule
    \end{tabularx}
    \caption{F1(\%) scores on three target tasks. Performance of our EG method using vanilla MS loss without external knowledge is reported as \textit{EG(MS)}. All the reported scores are averaged over 3 different random seeds.}
    \label{tab:eg_ablation}
\end{table}
\endgroup


\subsection{Generalization Ability}

\begingroup
\begin{table}[htb]
    \centering
    \small
    \begin{tabularx}{\linewidth}{>{\centering\arraybackslash\hsize=1.2\hsize}X>{\centering\arraybackslash\hsize=1.1\hsize}X>{\centering\arraybackslash\hsize=.9\hsize}X>{\centering\arraybackslash\hsize=.9\hsize}X>{\centering\arraybackslash\hsize=.9\hsize}X}
    \toprule
    \textbf{Event Annotator} & \textbf{Target Tasks} & \textbf{\textit{CHEMD}} & \textbf{\textit{BC5CDR}} & \textbf{\textit{DrugProt}} \\
    \toprule
    \multirow{2}{*}{Gold-std} & Concat & \underline{47.03} & 78.89 & 67.49 \\
    & SentEnc & \underline{46.91}  & \underline{79.76} & 67.93 \\
    \midrule
    \multirow{2}{*}{Auto-sys} & Concat & 46.06 & \underline{79.49} & \underline{68.31} \\
    & SentEnc & 46.11 & 79.39 & \underline{68.04} \\
    \bottomrule
    \end{tabularx}
    \caption{F1(\%) scores of our proposed EG methods based on human/machine annotated events. We highlight better scores between Gold-std and Auto-sys annotators under each setting with underlines. All the reported scores are averaged over 3 different random seeds.}
    \label{tab:eg_gen}
\end{table}
\endgroup

To alleviate the reliance on gold-standard event annotations, which may be hard to obtain, we generate the event annotations using DeepEventMine \cite{DeepEventMine}. Table~\ref{tab:eg_gen} reports the performance of the EG methods. We see that the performance is comparable to that of the gold-standard annotations. We also observe that in the DrugProt dataset, the performance with automatic annotations is better than that of the gold-standard annotations, suggesting that the human annotations are low quality and noisy.

\subsection{External Knowledge}

Table \ref{tab:eg_ablation} reports the performance of our EG method without external knowledge (i.e., event annotations), where simple MS loss replaces our RMS loss. The performance over three target tasks mirrors the \textit{Direct Transfer} setting, suggesting that the vanilla MS objective has minimal impact and the main improvement stems from the auxiliary data extracted from the event mentions.

\subsection{Pseudo Label Usage}
\label{appendix:ablation_exp_detail}

\noindent\paragraph{\textit{Pse-Augment}}
We augment annotations of target tasks with pseudo entities within the target corpus labeled as ``Out of Distribution (OOD)'' entities. Then, the model is trained with a single cross-entropy loss.
\noindent\paragraph{\textit{Pse-Classifer}}
We first use a pretrained model as a classifier separating pseudo entities and gold-standard annotated entities. We then predict with the former directly finetuned model and filter out entities labeled ``OOD'' by the classifier.

\begin{table}[htb]
    \centering
    \small
    \begin{tabularx}{\linewidth}{>{\centering\arraybackslash\hsize=1.8\hsize}X>{\centering\arraybackslash\hsize=1\hsize}X>{\centering\arraybackslash\hsize=0.6\hsize}X>{\centering\arraybackslash\hsize=0.6\hsize}X}
    \toprule
    \textbf{Dataset} & \textbf{Precision} & \textbf{Recall} & \textbf{F1} \\
    \toprule
    Few-shot & 42.73 & 51.69 & 46.77 \\
    Oracle & \textbf{73.56} & \textbf{80.03} & \textbf{76.66} \\
    \bottomrule
    \end{tabularx}
    \caption{Comparison of the target-only results for different training set sizes in CHEMDNER.}
    \label{tab:full_chemdner}
\end{table}
\subsection{Size of Target Set}
Since the training documents for the target domain are downsampled to roughly 10\% of the original corpus, we compare our downsampled training results (few-shot) with those trained on the entire target set (Oracle). We finetune the BERT model with adapters on the full target dataset CHEMDNER in Table~\ref{tab:full_chemdner}. The large gap indicates that the limited training data indeed severely hinders the model's learning of the task and justifies the need for transferring knowledge from a high-resource source domain.

\section{Visualization}
\label{appendix:visual}

\begin{figure*}[!htb]
    \centering
    \includegraphics[width=1.0\linewidth]{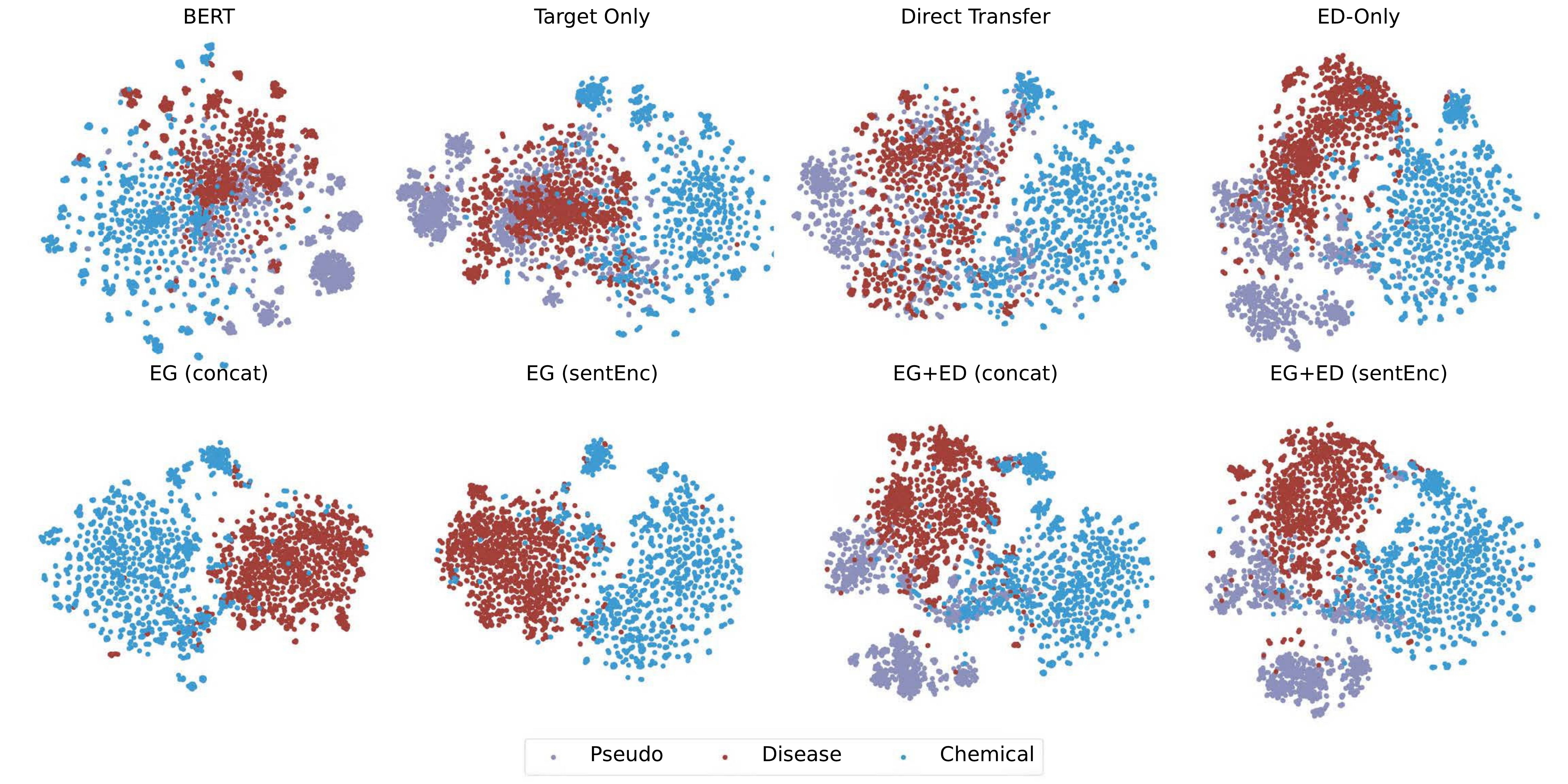}
    \caption{t-SNE visualization of entities in the test corpus of BC5CDR. \textit{Pseudo} is labeled by model pretrained on source task, \textit{Disease} and \textit{Chemical} are gold-standard annotations. \textit{BERT} represents vanilla BERT model without pretraining or finetuning, and all the settings are same as Main Results. }
    \label{fig:analysis_bc5cdr}
\end{figure*}

The t-SNE projection visualization of BC5CDR test entity representations is shown in Figure \ref{fig:analysis_bc5cdr}.

\section{Scientific Artifacts}

We list the license used in this paper: PathwayCuration (CC BY-SA 3.0), InfectiousDisease (CC BY-SA 3.0), CancerGenetics (CC BY-SA 3.0), CHEMDNER (CC BY 4.0), BC5CDR (CC BY 4.0), DrugProt (CC BY 4.0), Huggingface Transformers (Apache License 2.0), SapBERT (apache-2.0), S-PubMedBert-MS-MARCO-SCIFACT (apache-2.0), OpenAI (Terms of use\footnote{\url{openai.com/policies/terms-of-use}}). 
We follow the intended use of all the mentioned existing artifacts in this paper. 
\begingroup
\begin{table*}[!htb]
    \centering
    \small
    \begin{tabularx}{\linewidth}{>{\centering\arraybackslash\hsize=2.08\hsize}X>{\centering\arraybackslash\hsize=.88\hsize}X>{\centering\arraybackslash\hsize=.88\hsize}X>{\centering\arraybackslash\hsize=.88\hsize}X>{\centering\arraybackslash\hsize=.88\hsize}X>{\centering\arraybackslash\hsize=.88\hsize}X>{\centering\arraybackslash\hsize=.88\hsize}X>{\centering\arraybackslash\hsize=.88\hsize}X>{\centering\arraybackslash\hsize=.88\hsize}X>{\centering\arraybackslash\hsize=.88\hsize}X}
    \toprule
      \textbf{Evaluate Datasets} & \multicolumn{3}{c}{\textbf{\textit{CHEMDNER}}} & \multicolumn{3}{c}{\textbf{\textit{BC5CDR}}} & \multicolumn{3}{c}{\textbf{\textit{DrugProt}}} \\
      \textbf{Metrics} & \textbf{Precision} & \textbf{Recall} & \textbf{F1} & \textbf{Precision} & \textbf{Recall} & \textbf{F1} & \textbf{Precision} & \textbf{Recall} & \textbf{F1} \\
    \hline
    Few-shot (BERT) & 42.73 & 51.69 & 46.77 & 72.44 & 85.86 & 78.51 & 63.80 & 67.42 & 65.49 \\
    \hline
    \multicolumn{10}{c}{\textbf{\textit{Pathway Curation}}} \\
    \hline
    Direct Transfer & 42.51 & 49.70 & 45.82 & 66.79 & \textbf{88.79} & 76.22 & 60.94 & \textbf{68.70} & 64.52 \\
    EG(concat) & 44.75 & 51.78 & 48.00 & 71.49 & 87.04 & 78.45 & 64.45 & 66.39 & 65.31 \\
    EG(sentEnc) & 42.80 & 50.43 & 46.25 & 70.91 & 86.97 & 78.09 & 64.36 & 67.81 & 66.03 \\
    ED & \textbf{46.97} & \textbf{52.30} & \textbf{49.48} & 74.71 & 83.96  & 79.06 & \textbf{67.05} & 67.13 & 67.08 \\
    EG(concat)+ED & 43.44 & 51.54 & 47.11 & 72.34 & 86.09 & 78.61 & 66.59 & 66.13 & 66.32 \\
    EG(sentEnc)+ED & 43.34 & 51.73 & 47.16 & \textbf{75.66} & 86.14 & \textbf{80.55} & 66.15 & 68.32 & \textbf{67.21} \\
    \hline
    \multicolumn{10}{c}{\textbf{\textit{Infectious Diseases}}} \\
    \hline
    Direct Transfer & 41.57 & 48.30 & 44.65 & \textbf{74.92} & 82.51 & 78.53 & 61.96 & 67.81 & 64.75 \\
    EG(concat) & 45.26 & 50.65 & 47.77 & 72.28 & \textbf{87.41} & 79.12 & 63.76 & 65.25 & 64.43 \\
    EG(sentEnc) & \textbf{47.33} & 50.41 & \textbf{48.80} & 74.58 & 86.37 & \textbf{80.04} & 64.83 & 63.60 & 64.20 \\
    ED & 41.07 & 46.93 & 43.78 & 74.60 & 85.85 & 79.80 & 64.25 & \textbf{69.75} & 66.63 \\
    EG(concat)+ED & 43.37 & 50.26 & 46.43 & 73.83 & 85.48 & 79.23 & 62.31 & 66.87 & 64.47 \\
    EG(sentEnc)+ED & 41.50 & \textbf{52.19} & 46.18 & 74.70 & 85.95 & 79.93 & \textbf{65.06} & 69.11 & \textbf{67.02} \\
    \hline
    \multicolumn{10}{c}{\textbf{\textit{Cancer Genetics}}} \\
    \hline
    Direct Transfer & 45.22 & 51.80 & 48.27 & 72.37 & 86.56 & 78.82 & 62.94 & 69.61 & 66.07 \\
    EG(concat) & 40.59 & \textbf{53.19} & 46.01 & 71.44 & \textbf{87.22} & 78.53 & 65.31 & 66.67 & 65.98 \\
    EG(sentEnc) & 41.54 & 52.36 & 46.33 & 72.68 & 86.09 & 78.82 & \textbf{66.99} & 65.71 & 66.30 \\
    ED & \textbf{47.06} & 50.58 & \textbf{48.75} & \textbf{75.08} & 86.96 & \textbf{80.57} & 66.26 & \textbf{73.52} & \textbf{69.68} \\
    EG(concat)+ED & 45.16 & 51.72 & 48.16 & 73.37 & 85.06 & 78.76 & 65.60 & 67.67 & 66.64 \\
    EG(sentEnc)+ED & 41.89 & 50.17 & 45.61 & 73.81 & 86.03 & 79.45 & 65.59 & 68.47 & 66.99 \\
    \bottomrule
    \end{tabularx}
    \caption{Full evaluation results. Pathway Curation, Infectious Diseases and Cancer Genetics are set to be the source domain respectively. Experiment settings are the same as main results reported in Table \ref{tab:main}. All the reported scores are averaged over 3 different random seeds.}
    \label{tab:full_res}
\end{table*}
\endgroup

\begin{table*}
    \centering
    \footnotesize
    \small
    \begin{tabularx}{\linewidth}{>{\centering\arraybackslash\hsize=.3\hsize}X>{\arraybackslash\hsize=1.7\hsize}X}
    \toprule
    \textbf{Type} & \textbf{Template} \\
    \midrule
    Conversion & A specific trigger <Trigger> causes the transformation of a molecule <Theme> into another molecule <Product>. \\
    \midrule
    Phosphorylation & Indicated by the given trigger <Trigger>, a specific molecule <Theme> is modified by the addition of a phosphate group at a particular site <Site>, facilitated by another molecule <Cause>. \\
    \midrule
    Dephosphorylation & Indicated by the given trigger <Trigger>, a specific molecule <Theme> has a phosphate group removed from a particular site <Site>, facilitated by another molecule <Cause>. \\
    \midrule
    Acetylation & Indicated by the given trigger <Trigger>, a specific molecule <Theme> undergoes the addition of an acetyl group at a particular site <Site>, catalyzed by another molecule <Cause>. \\
    \midrule
    Deacetylation & Indicated by the given trigger <Trigger>, a specific molecule <Theme> has an acetyl group removed from a particular site <Site>, facilitated by another molecule <Cause>. \\
    \midrule
    Ubiquitination & Indicated by the given trigger <Trigger>, a specific molecule <Theme> is modified by the attachment of one or more ubiquitin molecules at a particular site, facilitated by another molecule <Cause>, often involving a simple chemical group <Site> as the site of attachment. \\
    \midrule
    Deubiquitination & Indicated by the given trigger <Trigger>, a specific molecule <Theme> has ubiquitin molecules removed from a particular site, facilitated by another molecule <Cause> involving a simple chemical group <Site> as the site of removal. \\
    \midrule
    Hydroxylation & Indicated by the given trigger <Trigger>, a specific molecule <Theme> undergoes the addition of a hydroxyl group at a particular site, catalyzed by another molecule <Cause> involving a simple chemical group <Site> as the site of attachment. \\
    \midrule
    Dehydroxylation & Indicated by the given trigger <Trigger>, a specific molecule <Theme> has a hydroxyl group removed from a particular site, facilitated by another molecule <Cause> involving a simple chemical group <Site> as the site of removal. \\
    \midrule
    Methylation & Indicated by the given trigger <Trigger>, a specific molecule <Theme> undergoes the addition of a methyl group at a particular site, facilitated by another molecule <Cause> involving a simple chemical group <Site> as the site of attachment. \\
    \midrule
    Demethylation & Indicated by the given trigger <Trigger>, a specific molecule <Theme> has a methyl group removed from a particular site, facilitated by another molecule <Cause> involving a simple chemical group <Site> as the site of removal. \\
    \midrule
    Localization & Indicated by the given trigger <Trigger>, a specific molecule <Theme> is directed to or away from a particular cellular component <AtLoc><FromLoc><ToLoc> or subcellular location within the cell. \\
    \midrule
    Transport & Indicated by the given trigger <Trigger>, a specific molecule <Theme> is moved or conveyed to or away from a particular cellular component <FromLoc><ToLoc> or subcellular location within the cell. \\
    \midrule
    Gene Expression & Indicated by the given trigger <Trigger>, genetic information from a gene <Theme> is used to produce a functional gene product, such as RNA or protein. \\
    \midrule
    Transcription & Indicated by the given trigger <Trigger>, genetic information from a gene <Theme> is transcribed into RNA, usually messenger RNA (mRNA), by RNA polymerase. \\
    \midrule
    Translation & Indicated by the given trigger <Trigger>, the genetic information carried by mRNA <Theme> is used to synthesize a protein by ribosomes in the cell. \\
    \midrule
    Degradation & Indicated by the given trigger <Trigger>, a specific molecule <Theme> undergoes breakdown or degradation into smaller components. \\
    \midrule
    Binding & Indicated by the given trigger <Trigger>, a specific molecule <Theme> interacts and forms a complex with another molecule(s) resulting in the product of a molecular complex <Product>. \\
    \bottomrule
    \end{tabularx}
    \caption{Templates for PathwayCuration Dataset.}
    \label{tab:full_temp_pc_0}
\end{table*}
\begin{table*}
    \centering
    \footnotesize
    \small
    \begin{tabularx}{\linewidth}{>{\centering\arraybackslash\hsize=.3\hsize}X>{\arraybackslash\hsize=1.7\hsize}X}
    \toprule
    \textbf{Type} & \textbf{Template} \\
    \midrule
    Binding & Indicated by the given trigger <Trigger>, a specific molecule <Theme> interacts and forms a complex with another molecule(s) resulting in the product of a molecular complex <Product>. \\
    \midrule
    Dissociation & Indicated by the given trigger <Trigger>, a specific complex <Theme> breaks apart, resulting in the release of individual molecules <Product> as products. \\
    \midrule
    Regulation & Indicated by the given trigger <Trigger>, an entity <Theme> is controlled or influenced by another entity <Cause> to achieve a specific biological effect or outcome. \\
    \midrule
    Positive Regulation & Indicated by the given trigger <Trigger>, an entity <Theme> is promoted or enhanced by another entity <Cause> to achieve a specific biological effect or outcome. \\
    \midrule
    Activation & Indicated by the given trigger <Trigger>, a specific molecule <Theme> is stimulated or facilitated by another entity <Cause> to increase its activity, function, or biological effect. \\
    \midrule
    Negative Regulation & Indicated by the given trigger <Trigger>, an entity <Theme> is inhibited or suppressed by another entity <Cause> to achieve a specific biological effect or outcome. \\
    \midrule
    Inactivation & Indicated by the given trigger <Trigger>, a specific molecule <Theme> is deactivated or rendered inactive by another entity <Cause>, leading to a reduction or cessation of its biological function. \\
    \midrule
    Pathway & Indicated by the given trigger <Trigger>, involving one or more molecules <Participant> that collaborate to accomplish a specific biological function or response. \\
    \bottomrule
    \end{tabularx}
    \caption{Continuation of templates for PathwayCuration Dataset.}
    \label{tab:full_temp_pc_1}
\end{table*}
\begin{table*}
    \centering
    \footnotesize
    \small
    \begin{tabularx}{\linewidth}{>{\centering\arraybackslash\hsize=.3\hsize}X>{\arraybackslash\hsize=1.7\hsize}X}
    \toprule
    \textbf{Type} & \textbf{Template} \\
    \midrule
    Gene Expression & Indicated by the given trigger <Trigger>, a specific protein or a group of genes <Theme> are involved in the transcription and translation of genetic information to produce functional gene products, such as RNA or protein. \\
    \midrule
    Transcription & Indicated by the given trigger <Trigger>, a specific protein or a group of genes <Theme> are involved in the synthesis of RNA from DNA template by RNA polymerase. \\
    \midrule
    Protein Catabolism & Indicated by the given trigger <Trigger>, a specific protein <Theme> is broken down or degraded into smaller peptide fragments or amino acids. \\
    \midrule
    Phosphorylation & Indicated by the given trigger <Trigger>, a specific protein <Theme> undergoes the addition of a phosphate group at a particular site <Site>, resulting in the modification of the protein's structure and function. \\
    \midrule
    Localization & Indicated by the given trigger <Trigger>, a specific core entity <Theme> is directed to or away one location <AtLoc><ToLoc> within the cell or organism. \\
    \midrule
    Binding & Indicated by the given trigger <Trigger>, a specific core entity <Theme> interacts and forms a connection with another entity <Site>, leading to the formation of a complex or association. \\
    \midrule
    Regulation & Indicated by the given trigger <Trigger>, a specific core entity or event <Theme> is controlled or influenced by another core entity or event <Cause> through interactions at specific sites  on molecules or entities <Site><CSite>, potentially resulting in modulation of biological processes. \\
    \midrule
    Positive Regulation & Indicated by the given trigger <Trigger>, a specific core entity or event <Theme> is promoted or enhanced by another core entity or event <Cause> through interactions at specific sites on molecules or entities <Site><CSite>, potentially resulting in an increase in the intensity or rate of a biological process. \\
    \midrule
    Negative Regulation & Indicated by the given trigger <Trigger>, a specific core entity or event <Theme> is inhibited or suppressed by another core entity or event <Cause> through interactions at specific sites on molecules or entities <Site><CSite>, potentially resulting in a decrease in the intensity or rate of a biological process. \\
    \midrule
    Process & Indicated by the given trigger <Trigger>, involving a core entity that collaborates to accomplish a specific biological function or response. \\  
    \bottomrule
    \end{tabularx}
    \caption{Templates for Infectious Diseases Dataset.}
    \label{tab:full_temp_id_0}
\end{table*}
\begin{table*}
    \centering
    \footnotesize
    \small
    \begin{tabularx}{\linewidth}{>{\centering\arraybackslash\hsize=.3\hsize}X>{\arraybackslash\hsize=1.7\hsize}X}
    \toprule
    \textbf{Type} & \textbf{Template} \\
    \midrule
    Development & Indicated by the given trigger <Trigger>, a specific anatomical or pathological entity <Theme> undergoes progressive changes or growth, leading to the formation of a more complex and specialized structure or condition over time. \\
    \midrule
    Blood Vessel Development & Indicated by the given trigger <Trigger>, blood vessels <Theme> undergo progressive changes or growth at a specific location <AtLoc>, leading to the formation and maturation of the vascular network. \\
    \midrule
    Growth & Indicated by the given trigger <Trigger>, a specific anatomical or pathological entity <Theme> undergoes an increase in size, quantity, or complexity over time. \\
    \midrule
    Death & Indicated by the given trigger <Trigger>, a specific anatomical or pathological entity <Theme> ceases to exhibit signs of life and undergoes irreversible loss of vital functions. \\
    \midrule
    Cell Death & Indicated by the given trigger <Trigger>, a specific cell <Theme> undergoes a series of events leading to its own demise, often through programmed cell death or other mechanisms. \\
    \midrule
    Breakdown & Indicated by the given trigger <Trigger>, a specific anatomical or pathological structure <Theme> disintegrates, decomposes, or undergoes degradation over time. \\
    \midrule
    Cell Proliferation & Indicated by the given trigger <Trigger>, a specific cell <Theme> undergoes rapid and controlled replication or division, leading to an increase in the number of daughter cells. \\
    \midrule
    Cell Division & Indicated by the given trigger <Trigger>, a specific cell <Theme> divides into two or more daughter cells through mitosis or meiosis. \\
    \midrule
    Cell Differentiation & Indicated by the given trigger <Trigger>, a specific cell <Theme> undergoes changes in gene expression and morphology to become specialized and acquire distinct functions at a specific anatomical or pathological location <AtLoc>. \\
    \midrule
    Remodeling & Indicated by the given trigger <Trigger>, a specific tissue <Theme> undergoes structural changes, reorganization, and modification in response to various stimuli or during growth and development. \\
    \midrule
    Reproduction & Indicated by the given trigger <Trigger>, a specific organism <Theme> produces offspring through sexual or asexual means, leading to the continuation of the species. \\
    \midrule
    Mutation & indicated by the given trigger <Trigger>, a specific gene, genome, or protein <Theme> undergoes a heritable change in its genetic sequence or structure at a particular site <Site>, potentially leading to alterations in its function or expression within a specific anatomical or pathological context <AtLoc>. \\
    \midrule
    Carcinogenesis & Indicated by the given trigger <Trigger>, a specific anatomical or pathological entity <Theme> undergoes a series of genetic and cellular changes at a specific anatomical or pathological location <AtLoc>, leading to the development of cancer. \\
    \midrule
    Cell Transformation & Indicated by the given trigger <Trigger>, a specific cell <Theme> undergoes changes in its phenotype, function, or behavior at a specific anatomical or pathological location <AtLoc>, often associated with the acquisition of abnormal or cancerous characteristics. \\
    \midrule
    Metastasis & Indicated by the given trigger <Trigger>, a specific anatomical or pathological entity <Theme> spreads and establishes secondary growths or lesions at a different anatomical or pathological location <ToLoc> from the primary tumor site. \\
    \midrule
    Infection & Indicated by the given trigger <Trigger>, an organism <Participant> invades and establishes itself in a specific anatomical or pathological site <Theme>, leading to a disease condition. \\
    \midrule
    Metabolism & Indicated by the given trigger <Trigger>, collective chemical reactions occur within an organism <Theme> involving the processing, transformation, and utilization of specific molecules to maintain cellular functions and energy requirements. \\
    \midrule
    Synthesis & Indicated by the given trigger <Trigger>, a specific simple chemical <Theme> is produced or created through chemical reactions or biological processes. \\
    \midrule
    Catabolism & Indicated by the given trigger <Trigger>, a specific molecule <Theme> undergoes chemical reactions or metabolic pathways to break down into simpler compounds, releasing energy in the process. \\
    \midrule
    Amino Acid Catabolism & Indicated by the given trigger <Trigger>, a specific amino acid <Theme> is broken down through metabolic pathways, leading to the release of energy and the generation of byproducts like ammonia and carbon compounds. \\
    \midrule
    Glycolysis & Indicated by the given trigger <Trigger>, a specific molecule <Theme> undergoes a series of enzymatic reactions, ultimately converting glucose into pyruvate and producing ATP and NADH as energy carriers. \\
    \bottomrule
    \end{tabularx}
    \caption{Templates for Cancer Genetics Dataset.}
    \label{tab:full_temp_cg_0}
\end{table*}
\begin{table*}
    \centering
    \footnotesize
    \small
    \begin{tabularx}{\linewidth}{>{\centering\arraybackslash\hsize=.3\hsize}X>{\arraybackslash\hsize=1.7\hsize}X}
    \toprule
    \textbf{Type} & \textbf{Template} \\
    \midrule
    Glycosylation & Indicated by the given trigger <Trigger>, a specific molecule <Theme> undergoes the addition of carbohydrate molecules (glycans) to specific sites, typically on proteins or lipids, leading to the formation of glycoproteins or glycolipids with diverse biological functions. \\
    \midrule
    Acetylation & Indicated by the given trigger <Trigger>, a specific molecule <Theme> undergoes the addition of an acetyl group at a particular site <Site>, catalyzed by another molecule <Cause>. \\
    \midrule
    Deacetylation & Indicated by the given trigger <Trigger>, a specific molecule <Theme> has an acetyl group removed from a particular site <Site>, facilitated by another molecule <Cause>. \\
    \midrule
    Ubiquitination & Indicated by the given trigger <Trigger>, a specific molecule <Theme> is modified by the attachment of one or more ubiquitin molecules at a particular site, facilitated by another molecule <Cause>, often involving a simple chemical group <Site> as the site of attachment. \\
    \midrule
    Deubiquitination & Indicated by the given trigger <Trigger>, a specific molecule <Theme> has ubiquitin molecules removed from a particular site, facilitated by another molecule <Cause> involving a simple chemical group <Site> as the site of removal. \\
    \midrule
    Gene Expression & Indicated by the given trigger <Trigger>, a specific gene, genome, or protein <Theme> is activated, leading to the production of RNA or protein and subsequent biological functions. \\
    \midrule
    Transcription & Indicated by the given trigger <Trigger>, genetic information from a specific gene, genome, or RNA molecule <Theme> is used as a template to produce complementary RNA (usually messenger RNA) by RNA polymerase. \\
    \midrule
    Translation & Indicated by the given trigger <Trigger>, the genetic information carried by messenger RNA <Theme> is used to synthesize a protein by ribosomes in the cell. \\
    \midrule
    Protein Processing & Indicated by the given trigger <Trigger>, the series of post-translational modifications, folding, and transportation of a specific gene, genome, or protein <Theme> in the cell to achieve its mature and functional form. \\
    \midrule
    Phosphorylation & Indicated by the given trigger <Trigger>, a specific molecule <Theme> undergoes the addition of a phosphate group at a particular site <Site> within a protein domain or region, often regulating the molecule's activity or function. \\
    \midrule
    Dephosphorylation & Indicated by the given trigger <Trigger>, a specific molecule <Theme> has a phosphate group removed from a particular site <Site> within a protein domain or region, often resulting in the modulation or termination of its activity or function. \\
    \midrule
    DNA Methylation & Indicated by the given trigger <Trigger>, a specific gene or genome <Theme> has a methyl group added to a particular site <Site> within a protein domain or region, often resulting in the regulation of gene expression and epigenetic modifications. \\
    \midrule
    DNA Demethylation & Indicated by the given trigger <Trigger>, a specific gene or genome <Theme> has a methyl group removed from a particular site <Site> within a protein domain or region, often resulting in the modulation of gene expression and epigenetic modifications. \\
    \midrule
    Pathway & Indicated by the given trigger <Trigger>, involving a specific molecule <Participant> that collaborates to accomplish a specific biological function or response. \\
    \midrule
    Binding & Indicated by the given trigger <Trigger>, a specific molecule <Theme> forms a physical association or interaction with another molecule or region <Site> on a protein or DNA, potentially leading to functional changes or regulatory effects. \\
    \midrule
    Dissociation & Indicated by the given trigger <Trigger>, a specific molecule <Theme> separates or detaches from another molecule or region <Site> on a protein or DNA, leading to the termination or disruption of their interaction or complex formation. \\
    \midrule
    Localization & Indicated by the given trigger <Trigger>, a specific molecule <Theme> is directed to or away from a particular cellular component <AtLoc><FromLoc><ToLoc> or subcellular location within the cell. \\
    \midrule
    Regulation & Indicated by the given trigger <Trigger>, an entity <Theme> is controlled or influenced by another entity <Cause> to achieve a specific biological effect or outcome. \\
    \midrule
    Positive Regulation & Indicated by the given trigger <Trigger>, an entity <Theme> is promoted or enhanced by another entity <Cause> to achieve a specific biological effect or outcome. \\
    \midrule
    Negative Regulation & Indicated by the given trigger <Trigger>, an entity <Theme> is inhibited or suppressed by another entity <Cause> to achieve a specific biological effect or outcome. \\
    \midrule
    Planned Process & Indicated by the given trigger <Trigger>, a specific entity <Theme> is involved or manipulated using an instrument <Instrument> for a predetermined outcome or purpose. \\
    \bottomrule
    \end{tabularx}
    \caption{Continuation of templates for Cancer Genetics Dataset.}
    \label{tab:full_temp_cg_1}
\end{table*}

\end{document}